\documentclass[conference]{IEEEtran}
\IEEEoverridecommandlockouts

\usepackage{cite}
\usepackage{amsmath,amssymb,amsfonts}
\usepackage{algorithmic}
\usepackage{graphicx}
\usepackage{textcomp}
\usepackage{xcolor}
\usepackage{booktabs}
\usepackage{multirow}
\usepackage{url}

\IEEEaftertitletext{\vspace{-20pt}}

\begin{document}

\title{Realistic Market Impact Modeling for Reinforcement Learning Trading Environments}

\author{\IEEEauthorblockN{Lucas Riera Abbade}
\IEEEauthorblockA{\textit{Escola Polit\'{e}cnica} \\
\textit{Universidade de S\~{a}o Paulo}\\
lucasabbade@usp.br}
\and
\IEEEauthorblockN{Anna Helena Reali Costa}
\IEEEauthorblockA{\textit{Escola Polit\'{e}cnica} \\
\textit{Universidade de S\~{a}o Paulo}\\
anna.reali@usp.br}
}

\maketitle

\begin{abstract}
Reinforcement learning (RL) has shown promise for trading, yet most open-source backtesting environments assume negligible or fixed transaction costs, which can cause agents to learn undesirable trading behaviors that fail on real-world execution.
We introduce a suite of three Gymnasium-compatible trading environments---MACE (Market-Adjusted Cost Execution) stock trading, margin trading, and portfolio optimization---that integrate nonlinear market impact models grounded in the Almgren--Chriss (AC) framework and the empirically validated square-root impact law.
Each environment shares a unified infrastructure: pluggable cost models, permanent impact tracking with exponential decay, and comprehensive trade-level logging.
We evaluate five DRL algorithms (A2C, PPO, DDPG, SAC, TD3) on the NASDAQ~100 universe, comparing a fixed 10\,bps baseline cost model against the AC model with Optuna-optimized hyperparameters.
Our results show that (i)~the cost model materially affects both absolute performance and the relative ranking of algorithms across all three environments---in stock trading, optimized PPO achieves the best OOS result (34\% total return, Sharpe 1.06) under the baseline but drops to 25\% under AC, while TD3 improves from 24\% to 30\%; in portfolio optimization, TD3 under AC achieves 32\% total return (best overall) while the same algorithm under baseline yields only 26\% (worst among agents); (ii)~the AC model produces dramatically different trading behavior---non-optimized TD3 in stock trading sees daily costs drop from \$200k to \$8k when switching to AC, with turnover falling from 19\% to 1\%; (iii)~hyperparameter optimization (HPO) is essential for constraining pathological trading---without HPO, agents exhibit monotonically increasing participation rates across epochs (e.g., SAC stock trading costs drop 82\% with HPO); and (iv)~algorithm choice interacts strongly with the cost model in environment-specific ways---in the margin environment, DDPG's OOS Sharpe jumps from $-$2.1 to 0.3 under AC while SAC's drops from $-$0.5 to $-$1.2.
We release the full environment suite as an open-source extension to FinRL-Meta.
\end{abstract}


\section{Introduction}\label{sec:intro}

Reinforcement learning (RL) has emerged as a powerful paradigm for quantitative trading and portfolio optimization, capable of discovering dynamic strategies from market data~\cite{sun2023}.
However, a persistent reality gap exists between simulated and live performance.
Many prior studies train agents in idealized environments that neglect the market impact of their trades---the permanent and temporary price changes caused by large orders consuming finite liquidity~\cite{almgren2001}.
A common simplification is to assume a fixed transaction cost of 10 basis points (bps), which does not scale with trade size, volatility, or available volume.
This oversight allows agents to trade with unrealistic frequency, producing inflated backtest results~\cite{detzel2023}.

We present a suite of three Gymnasium-compatible environments~\cite{brockman2016} that embed a nonlinear market impact model, calibrated using the Almgren--Chriss (AC) framework~\cite{almgren2001} and the square-root impact law~\cite{Toth2011}.
The environments extend the FinRL-Meta library~\cite{liu2022} and are designed for multi-asset trading.
All three environments share a common infrastructure:
\begin{enumerate}
  \item \textbf{Pluggable impact models with permanent impact tracking}: Almgren-Chriss, square-root, Obizhaeva--Wang~\cite{obizhaeva2013}, and a 10\,bps baseline are available, others can be easily added using the same interface.
  \item \textbf{Trade-level logging}: POV (order percentage of volume), turnover percentile, and per-stock permanent impact for detailed backtest cost analysis.
  \item \textbf{Report generator}: Detailed report generator to visualize returns, sharpe, drawdowns, epoch evolution metrics, and trading costs.
\end{enumerate}

The three environments represent distinct RL formulations of portfolio management:

\smallskip\noindent\textbf{MACE Stock Trading} (long-only): the Market-Adjusted Cost Execution (MACE) environment is our primary contribution. The agent outputs a continuous signal $a_i \in [-1,1]$ per stock, converted to share quantities respecting position and volume limits. Uses a DSR reward~\cite{moody1998} with drawdown penalty and online observation normalization (running-mean/variance with serializable state for train-to-test transfer).

\smallskip\noindent\textbf{Margin Trading} (long/short with leverage): adapted from Gu et al.~\cite{gu2023margin}, who use the Dow Jones 30. We expand the universe to the NASDAQ~100 to test market impact on a larger set of stocks. Uses a profit $+$ rolling Sharpe reward.

\smallskip\noindent\textbf{Portfolio Optimization (POE)} (long-only): adapted from Costa \& Costa~\cite{costa2023poe}, who use a small set of Brazilian equities. We again expand to the NASDAQ~100 for the same reason. Uses a log-return reward.

Our contributions are as follows.
\begin{itemize}
  \item An open-source, unified environment suite with realistic market impact for RL trading research.
  \item A systematic comparison of five DRL algorithms (A2C, PPO, DDPG, SAC, TD3) across three environment formulations under baseline (10 bps) vs.\ AC impact models, with Optuna-based hyperparameter optimization.
  \item Evidence that the nonlinear AC cost signal produces qualitatively different training dynamics---enabling OOS convergence, constraining over-trading, and altering the relative ranking of algorithms compared to flat-fee baselines.
  \item Analysis showing that hyperparameter optimization is critical not only for OOS performance but for preventing pathological execution behavior such as unbounded growth in participation rates.
\end{itemize}

\section{Market Impact Models}\label{sec:impact}

When a trader submits an order, executing against available liquidity moves the price against them, generating execution costs commonly referred to as \emph{market impact}.
The optimal-execution literature, notably the framework of Almgren and Chriss~\cite{almgren2001}, decomposes these costs into a \emph{temporary} component (the instantaneous cost of demanding liquidity, which reverts once the order is complete) and a \emph{permanent} component (a lasting price shift reflecting the information content of the trade); the phenomenon has deeper roots in market-microstructure models of price formation such as Kyle's model of informed trading~\cite{kyle1985}.
The magnitude of both components depends on order size relative to available volume, asset volatility, and execution speed---none of which are captured by a flat basis-point fee~\cite{bouchaud2009}.
Below we describe the two models used in our environments: the empirically validated square-root law (Section~\ref{sec:sqrt}) and the Almgren--Chriss cost decomposition (Section~\ref{sec:ac}), followed by our treatment of permanent impact decay (Section~\ref{sec:decay}).

\subsection{The Square-Root Impact Law}\label{sec:sqrt}
Empirical studies document that the price impact of a metaorder scales as the square root of the fraction of average daily volume (ADV) traded~\cite{Toth2011,bouchaud2018trades,almgren2005}:
\begin{equation}\label{eq:sqrt}
  I(Q) = Y \cdot \sigma \cdot \sqrt{\frac{Q}{V}}\,,
\end{equation}
where $I(Q)$ is the expected price change, expressed as a return, measured at end of execution, $Q$ is the (unsigned) metaorder size, $V$ is the average daily volume, $\sigma$ is daily return volatility, and $Y$ is an empirical prefactor typically calibrated in the range $0.5$--$1.0$ depending on the market and dataset~\cite{bouchaud2018trades}.
This reduced-form expression averages over execution styles and participation rates; richer models incorporate duration and schedule effects~\cite{bacry2015,bouchaud2018trades}.
After execution, impact partially decays; fair-pricing models~\cite{farmer2013} predict a long-run level of $\sim\!\tfrac{2}{3}$ of peak impact, though empirical estimates vary with horizon and the decomposition of informational versus mechanical components~\cite{brokmann2015,bacry2015}.

\subsection{Almgren--Chriss Cost Decomposition}\label{sec:ac}
Following the linear impact framework of~\cite{almgren2001,almgren2005}, we decompose the execution cost of a single-period trade of $x$ shares at price $P$, with daily volatility $\sigma$ and average daily volume $V$, into three components:
\begin{align}
  C_\text{perm}   &= \tfrac{1}{2}\,\alpha\,\sigma\,
                      \bigl(x/V\bigr)\,|x|\,P\,,
                      \label{eq:ac-perm}\\
  C_\text{spread} &= \varepsilon\,|x|\,P\,,
                      \label{eq:ac-spread}\\
  C_\text{temp}   &= \beta\,\sigma\,
                      \bigl(x/V\bigr)\,|x|\,P\,,
                      \label{eq:ac-temp}
\end{align}
where $\alpha$ controls permanent impact, $\varepsilon$ is the half-spread (default $5$\,bps), and $\beta$ controls temporary depth-depletion cost. The permanent price shift is $\Delta P = \alpha\,\sigma\,(x/V)\,P$.

\subsection{Permanent Impact Decay}\label{sec:decay}
Real markets absorb the information content of trades over time.
We model this with exponential decay: $\Delta P_t = \Delta P_{t-1}(1 - \lambda)$ each trading day, with half-life $\tau_{1/2}$ days so $\lambda = 1 - 2^{-1/\tau_{1/2}}$.
Default $\tau_{1/2} = 5$ days for large-cap equities~\cite{bouchaud2009}.

\section{Environment Design}\label{sec:environment}

All three environments share a common Gymnasium interface: an observation vector encoding market state and portfolio positions, a continuous action space mapped to trades, and a reward signal that balances risk-adjusted return against execution costs.
The MACE (Market-Adjusted Cost Execution) stock trading environment is our primary contribution and is described in detail below. The margin trading and portfolio optimization environments preserve the state representation, action semantics, and reward functions defined by their original authors~\cite{gu2023margin,costa2023poe}; our modifications to those environments are limited to integrating the pluggable impact model interface, permanent impact tracking, and trade-level logging described in Section~\ref{sec:intro}.

\subsection{State Space}
The observation is a flat vector composed of:
\begin{itemize}
  \item Cash (or margin features) as a fraction of total equity.
  \item One-day log returns: $\log(P^{\text{adj}}_t / P^{\text{adj}}_{t-1})$.
  \item Position value as a fraction of total equity.
  \item Technical indicators (MACD, RSI, CCI, etc.), scaled by $2^{-7}$.
  \item Holdings relative to 20-day ADV: $h_i / \text{ADV}_{20,i}$.
  \item \emph{Optional}: permanent impact in basis points, cooldown counters, risk-free rate.
\end{itemize}
Observations are optionally standardized via online running-mean/variance normalization with serializable state for train$\to$test transfer.

\subsection{Action Space and Trade Execution}

\paragraph{Stock and Margin Trading}
Actions $a_i \in [-1,1]$ per stock.
Let $\bar{n}_i = \lfloor \phi\, V / P_i \rfloor$ denote the maximum position size in shares, where $V$ is portfolio value, $P_i$ the current price, and $\phi$ the per-stock exposure limit (starting parameter).
The desired trade is $\Delta_i = a_i \times \bar{n}_i$; the resulting position is clamped to $[-\bar{n}_i,\;\bar{n}_i]$ and the trade is further clipped to a maximum fraction of daily volume.
Sells are processed before buys to free cash.

\paragraph{Portfolio Optimization}
Actions are raw logits for $N+1$ slots (cash plus $N$ stocks), passed through a softmax to produce target weights $w_i$.
The environment computes the share-level rebalancing trades required to reach those targets, clips each trade to a maximum fraction of daily volume, and applies impact costs.

\subsection{Reward Function}
We use the Differential Sharpe Ratio (DSR)~\cite{moody1998} as the primary reward signal for the MACE stock trading environment.
The margin trader and portfolio optimization environments retain the reward functions defined by their respective authors~\cite{gu2023margin,costa2023poe}.

Let $r_t = V_t / V_{t-1} - 1$ be the one-step portfolio return. The DSR update is:
\begin{align}
  \mu_t &= (1-\alpha)\mu_{t-1} + \alpha\,r_t\,,\quad
  m^2_t = (1-\alpha)m^2_{t-1} + \alpha\,r_t^2\,,\\
  \text{DSR}_t &= \frac{r_t - \mu_{t-1}}{\hat\sigma_t} - \frac{1}{2}\,\text{SR}_{t-1}\left(\frac{r_t - \mu_{t-1}}{\hat\sigma_t}\right)^2\,,
\end{align}
where $\alpha = 1/H$ with horizon $H$, $\hat\sigma_t = \sqrt{\max(m^2_t - \mu_t^2,\,\epsilon)}$, and $\text{SR}_{t-1} = \mu_{t-1}/\hat\sigma_{t-1}$.
The drawdown penalty augments the DSR:
\begin{equation}
  R_t = \bigl(\text{DSR}_t - \eta_{\text{dd}}\,(\Delta\text{DD}_t)^2\bigr) \times s\,,
\end{equation}
where $\Delta\text{DD}_t = \max(0,\,\text{DD}_t - \text{DD}_{t-1})$ and $s$ is a scaling factor.

\begin{figure*}[!t]
\centering
\includegraphics[width=0.85\textwidth]{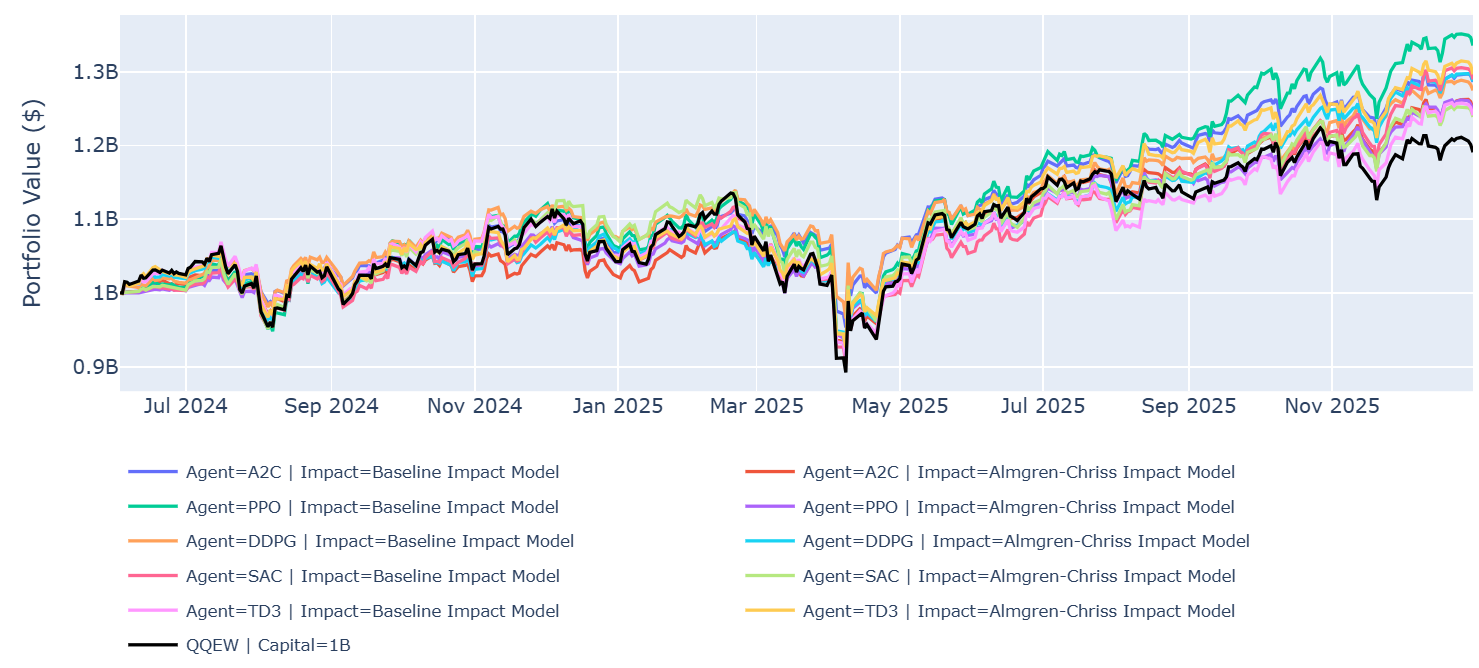}
\caption{OOS total return---MACE stock trading, all five agents under baseline vs.\ AC impact, optimized params. Black line: QQEW benchmark (19\%).}
\label{fig:stock_oos_return_bar}
\end{figure*}

\section{Experimental Setup}\label{sec:methodology}

We evaluate five DRL algorithms on the NASDAQ~100 universe under two cost models (flat 10\,bps baseline and Almgren--Chriss), with and without Optuna hyperparameter optimization.
This section details the data, algorithms, HPO pipeline, and comparison protocol.

\subsection{Data and Universe}
We use the NASDAQ~100 universe with daily open, high, low, close, and volume data from January 2010 to January 2026, split 90/10 for training and out-of-sample (OOS) testing.
HPO is performed on the data up until January 2025, with 2025 data being reserved for true OOS and final evaluation.

The benchmark is the QQEW ETF (equal-weighted NASDAQ~100), chosen because the agent's per-stock cap of 2\% makes an equal-weighted benchmark more comparable than the cap-weighted QQQ.
Technical indicators include MACD, RSI, CCI, Bollinger Bands, and moving averages from the FinRL indicator set.

Note that we use a static NASDAQ~100 composition from circa 2021 that's available in FinRL. While this introduces mild survivorship bias, the old list actually \emph{mitigates} the bias in the OOS period relative to using the current composition, since stocks that were subsequently removed are included in training.

\subsection{DRL Algorithms}
We evaluate five algorithms from Stable-Baselines3~\cite{raffin2021}: A2C, PPO, DDPG, SAC, and TD3.
Each is trained epoch-by-epoch (one epoch $=$ one pass through the training period) with per-epoch OOS evaluation to track overfitting.

\subsection{Hyperparameter Optimization}
We use Optuna~\cite{akiba2019} with TPE sampling and median pruning for HPO.
The search space covers both environment parameters (reward scaling, DSR horizon, drawdown penalty weight, observation features, normalization) and algorithm-specific parameters (learning rate, network architecture, entropy coefficient, etc.).
The objective is the best OOS annualized Sharpe ratio across epochs.

\subsection{Comparison Protocol}
For the stock trading and POE environments we run a combination of the 5 agents with (i)~baseline (10 bps) vs Almgren-Chriss cost model, and (ii)~default hyperparameters vs optimized hyperparameters, totaling 20 backtests.
For the margin trader environment we stick to the configuration in \cite{gu2023margin}, where they didn't include TD3 to the comparison and use the hyperparameters as described in their work, only switching the cost models, totaling 8 backtests.

\section{Results}\label{sec:results}

We present results for each environment in turn.
The stock trading (Section~\ref{sec:stock}) and portfolio optimization (Section~\ref{sec:poe}) environments share the same comparison grid---five agents, two cost models, default and optimized hyperparameters---but reveal different dynamics: stock trading shows the strongest HPO effects and the most extreme cost differences, while POE produces the highest absolute returns and the clearest evidence that the AC cost signal improves training convergence.
The margin trading environment (Section~\ref{sec:margin}) uses default parameters only (following the original paper's configuration) and isolates the cost model effect without HPO.

\subsection{MACE Stock Trading: Five-Agent Comparison}\label{sec:stock}

All configurations outperform the QQEW benchmark (19\% OOS total return) except non-optimized TD3 under AC.
The best result is optimized PPO under the baseline cost model: 34\% total return and 1.06 Sharpe, switching to the AC cost model reduces total return to 25\% (Sharpe 1.03), but trading costs drop 55\%, average order POV from 0.18 to 0.14, max drawdown from $-$20\% to $-$16\%, turnover from 1.3\% to 1.1\%, and volatility from 19\% to 15\%---a substantially less risky portfolio.
TD3 shows the opposite pattern: AC \emph{improves} total returns from 24\% to 30\% and Sharpe from 0.9 to 1.1, while turnover drops from 4\% to 3\%, costs fall 13\%, and POV decreases from 0.36 to 0.20. DDPG also improves modestly under AC; A2C and SAC are slightly worse.

\begin{figure}[t]
\centering
\includegraphics[width=\columnwidth]{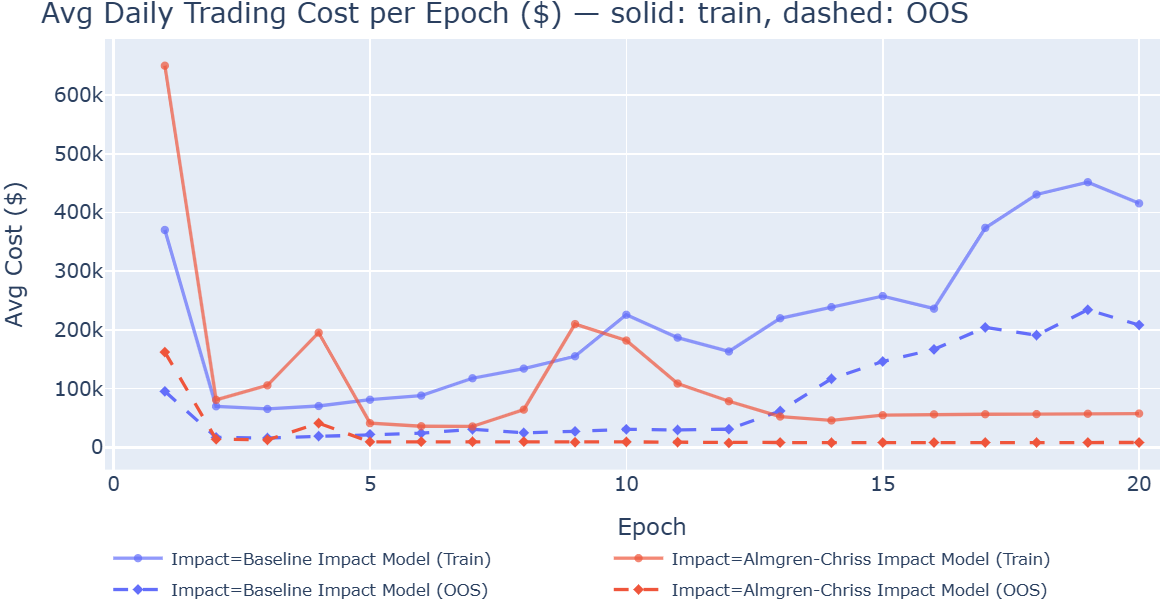}
\caption{Non-optimized TD3 trading costs---MACE stock trading, baseline vs.\ AC cost model.}
\label{fig:stock_td3_cost_extreme}
\end{figure}

HPO dramatically reduces trading activity.
Optimized SAC cuts turnover from 5\% to 2\% on AC and trading costs by 82\% (POV from 0.46 to 0.18); DDPG shows a similar pattern.
All optimized agents place heavy emphasis on cost reduction, sometimes at the expense of raw returns.
The most extreme case is non-optimized TD3 under the baseline: average order POV of 1\%, daily turnover of 19\%, and average daily trading cost of \$200k. The same agent under AC achieves POV of 0.2\%, turnover of 1\%, and daily cost of \$8k---a 96\% reduction that only moderately affects total returns (22\% vs.\ 18\%) because lower volatility (17\% to 12\%) preserves Sharpe (0.8 to 0.9) (Fig.~\ref{fig:stock_td3_cost_extreme}).

\begin{figure}[t]
\centering
\includegraphics[width=\columnwidth]{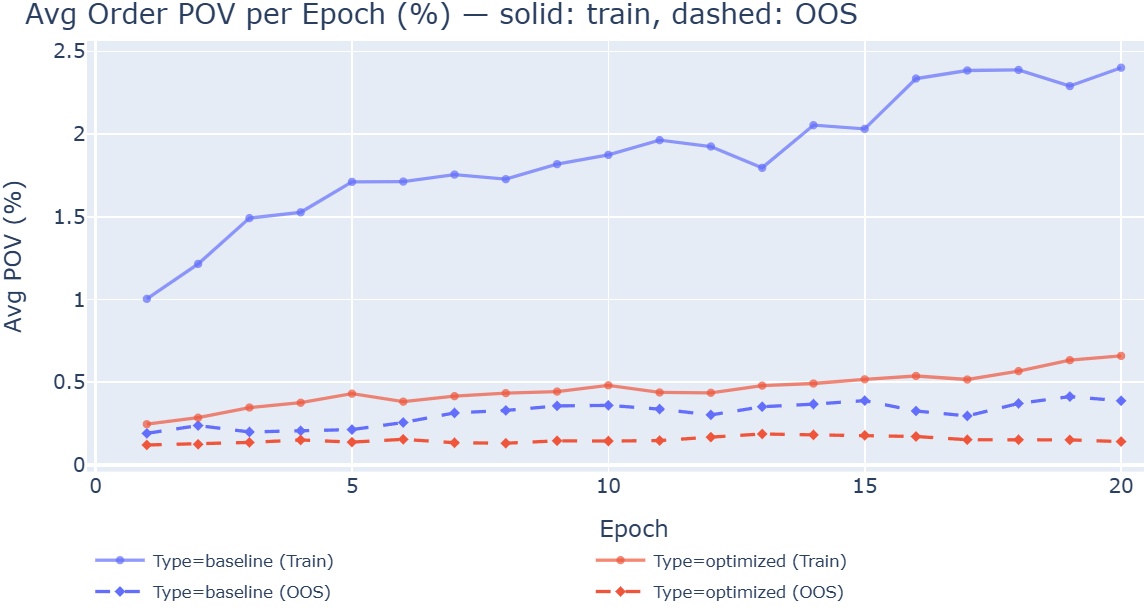}
\caption{Average order POV per epoch---MACE stock trading, PPO, optimized vs.\ non-optimized (AC impact model).}
\label{fig:stock_epoch_pov}
\end{figure}

Per-epoch analysis reveals that PPO, SAC, and A2C exhibit rising POV and turnover over training epochs---a tendency that HPO substantially mitigates. Optimized PPO's IS POV on AC rises from 0.25\% to 0.65\%, vs.\ 1\% to 2.4\% for non-optimized (Fig.~\ref{fig:stock_epoch_pov}).
Turnover follows the same pattern: non-optimized PPO 2.2\% to 7.9\% vs.\ optimized 0.54\% to 1.47\%. SAC shows similar dynamics (optimized turnover 1.12\% to 3.85\%, POV 0.72\% to 1.56\%; non-optimized stable at $\sim$8\% turnover and $\sim$2.5\% POV).
A2C exhibits the same behavior to a lesser extent.

\subsection{Margin Trading: Four-Agent Comparison}\label{sec:margin}

\begin{figure}[t]
\centering
\includegraphics[width=\columnwidth]{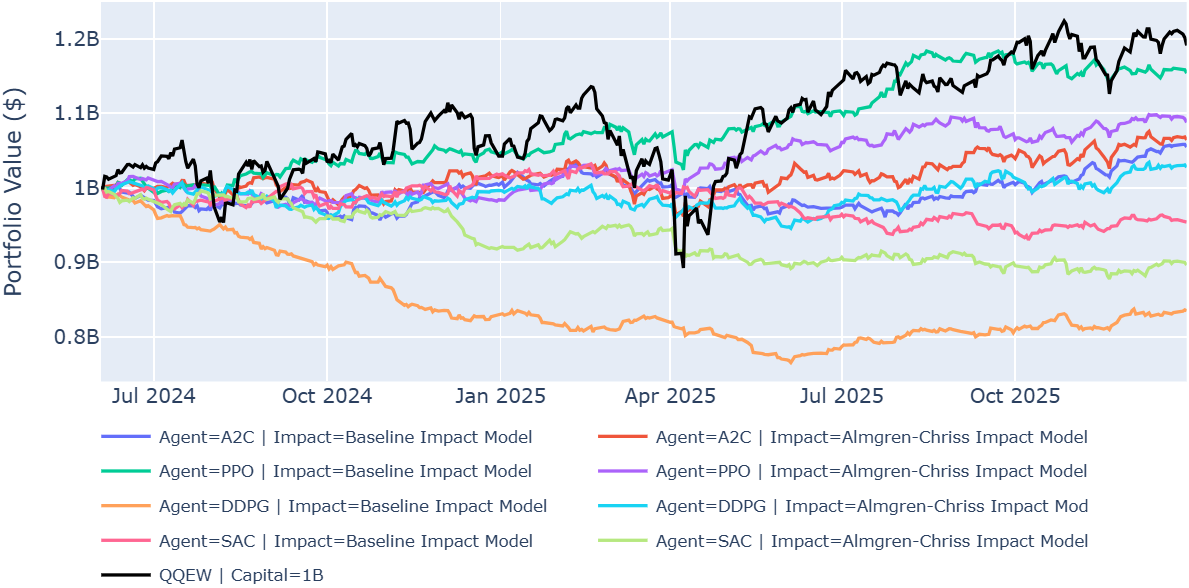}
\caption{OOS portfolio value---margin trading, A2C/PPO/DDPG/SAC, baseline vs.\ AC impact model.}
\label{fig:margin_portfolio_value}
\end{figure}

All agents underperform the QQEW benchmark OOS, although A2C and PPO outperform IS, suggesting overfitting that HPO could mitigate. The AC model's effect varies significantly by agent.
A2C total returns are 5.5\% (baseline) and 6.4\% (AC) OOS, with IS annualized returns of 13\% and 17.5\% respectively---the IS--OOS gap is large but the AC model produces slightly better OOS performance. PPO achieves 15\% OOS total return under the baseline but only 9\% under AC, despite similar turnover ($\sim$3\%) and 40\% lower trading costs on AC.
DDPG and SAC are the weakest performers, and the cost model makes a large difference in opposite directions: DDPG improves dramatically under AC (OOS Sharpe from $-$2.1 to 0.3, annualized return from $-$11\% to 2\%, max drawdown from $-$23\% to $-$6\%) with similar trading costs and POV, while SAC degrades under AC (OOS Sharpe from $-$0.5 to $-$1.2, annualized return from $-$3\% to $-$7\%), with lower POV (0.26 to 0.15) but $\sim$10\% higher trading costs (Fig.~\ref{fig:margin_portfolio_value}).

\begin{figure}[t]
\centering
\includegraphics[width=\columnwidth]{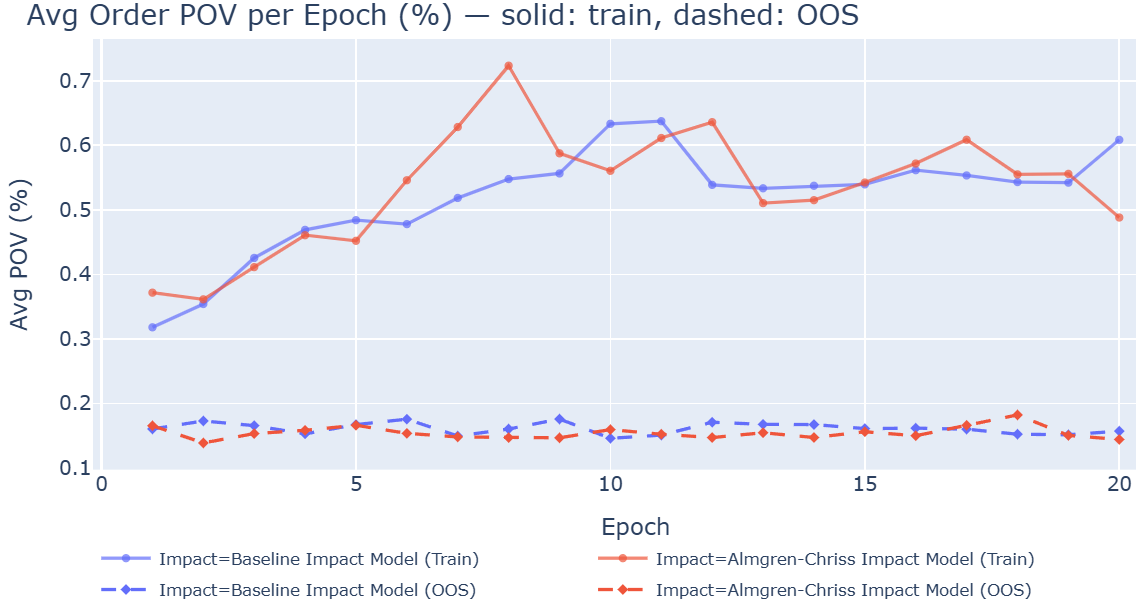}
\caption{PPO average order POV per epoch---margin trading.}
\label{fig:margin_ppo_epoch_pov}
\end{figure}

Training dynamics differ across algorithms.
A2C behaves similarly under both cost models, with OOS Sharpe and return peaking around epoch~15 and degrading afterward; other metrics remain stable. PPO exhibits IS--OOS divergence---IS return rises slightly while OOS declines over epochs, more so under AC than baseline.
PPO's average trading costs and POV spike by epoch~10 (IS POV rising from $\sim$0.37 to $\sim$0.6) before partially receding to $\sim$0.5 (Fig.~\ref{fig:margin_ppo_epoch_pov}).
DDPG and SAC show no evolution across epochs for any metric, consistent with early convergence to a near-deterministic policy, which could be improved with HPO.

\subsection{Portfolio Optimization: Five-Agent Comparison}\label{sec:poe}

\begin{figure}[t]
\centering
\includegraphics[width=\columnwidth]{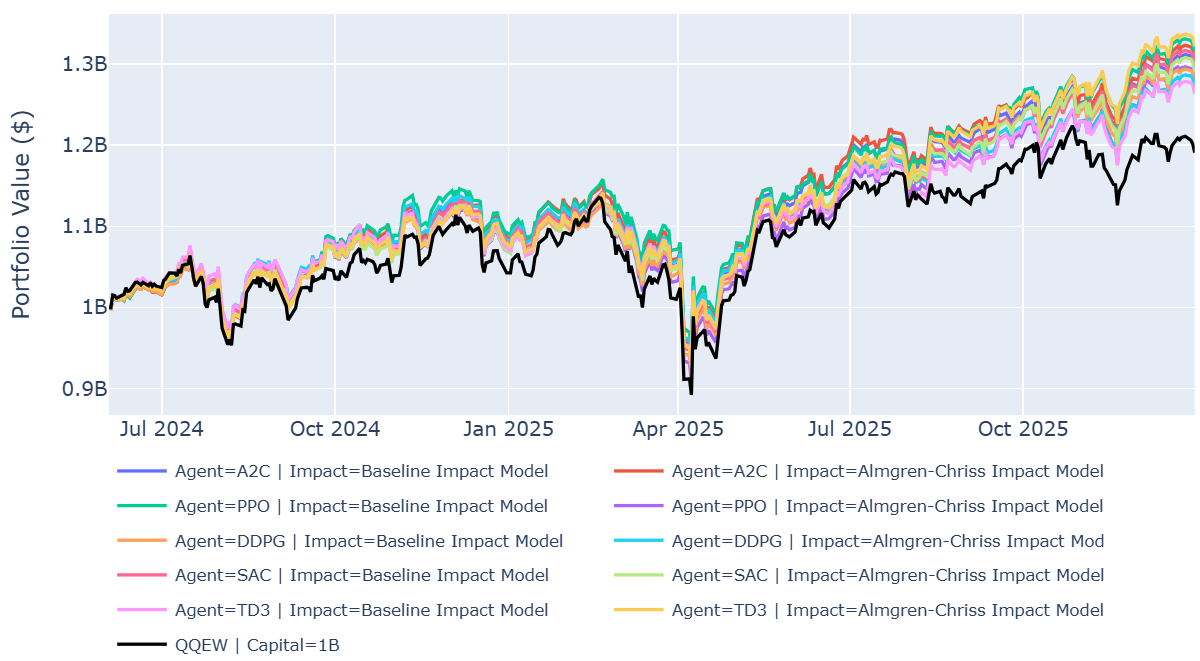}
\caption{OOS total return---POE, all five agents under 10\,bps baseline vs.\ AC impact.}
\label{fig:poe_oos_return_bar}
\end{figure}

All agents outperform the benchmark.
TD3 exhibits the widest spread across cost models: with the AC model it achieves 32\% total return---the highest OOS result of any configuration---while with the 10\,bps baseline it returns only 26\%, the lowest among all agents. A2C obtains $\sim$30\% total return under both models (slightly higher with AC); SAC reaches 30\% (baseline) and 29\% (AC); DDPG is comparable at 28\% (baseline) and 27\% (AC).
PPO is the only agent whose AC performance is meaningfully \emph{worse} than its baseline: 28\% vs.\ 31\%. The annualized Sharpe ratio improves from 0.9 to 1.1 for TD3 when switching from the baseline to AC; other agents remain in the 0.9--1.0 range.
TD3 total trading cost decreases by $\sim$15\% under AC relative to the baseline, while the remaining agents see $\sim$30\% lower costs (Fig.~\ref{fig:poe_oos_return_bar}).

\begin{figure}[t]
\centering
\includegraphics[width=\columnwidth]{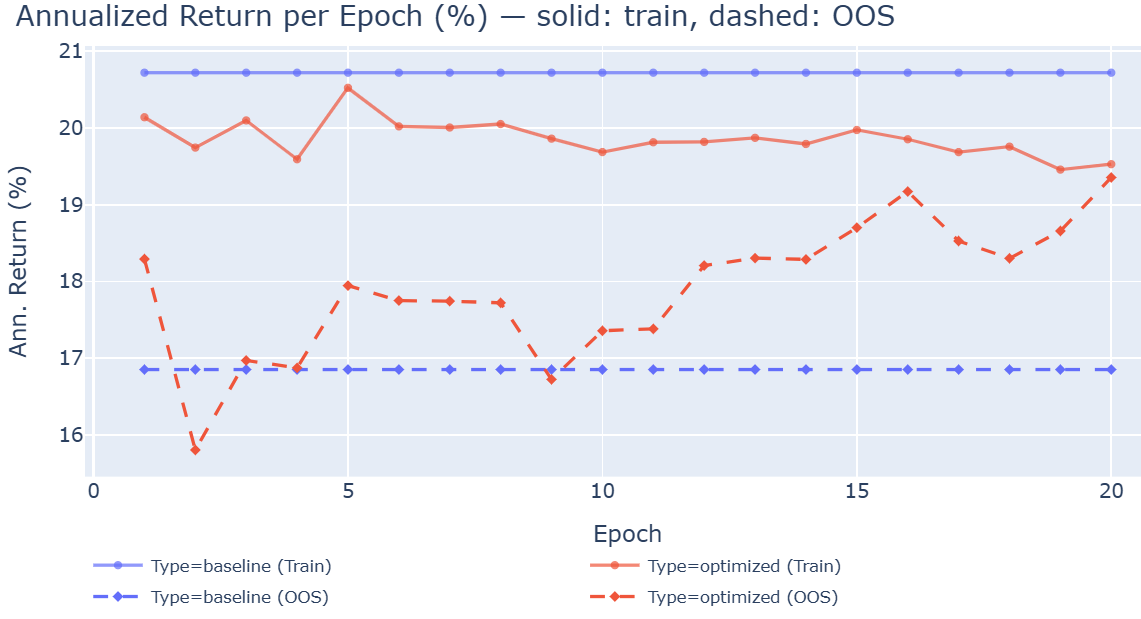}
\caption{TD3 return and Sharpe per epoch---POE, baseline vs.\ optimized parameters.}
\label{fig:poe_td3_epoch}
\end{figure}

Training-epoch dynamics under the AC impact model reveal clear benefits of HPO.
For TD3, the baseline-parameter agent has higher IS return than the optimized agent, yet lower OOS return; both IS and OOS metrics remain flat across epochs under baseline parameters.
The optimized TD3 exhibits genuine convergence: IS return decreases gradually while OOS return improves substantially, and the same pattern holds for Sharpe (Fig.~\ref{fig:poe_td3_epoch}).
Drawdown is unaffected by HPO. Turnover and average order POV increase over epochs for the optimized agent (final OOS POV: 0.13 vs.\ 0.10 for baseline; IS POV rises to 0.68 for optimized vs.\ 0.37 flat for baseline) but remain constant for the baseline.
A2C shows a similar but less pronounced convergence pattern.
PPO peaks at epoch~10 and degrades afterward; critically, baseline-parameter PPO exhibits monotonically increasing POV and trading costs at every epoch (IS POV reaching 1.39 vs.\ 0.36 for optimized; OOS 0.16 vs.\ 0.11), confirming that HPO is essential for constraining PPO's tendency toward over-trading.
SAC returns and Sharpe are highly unstable across epochs for both configurations, oscillating substantially from epoch to epoch.

\begin{figure}[t]
\centering
\includegraphics[width=\columnwidth]{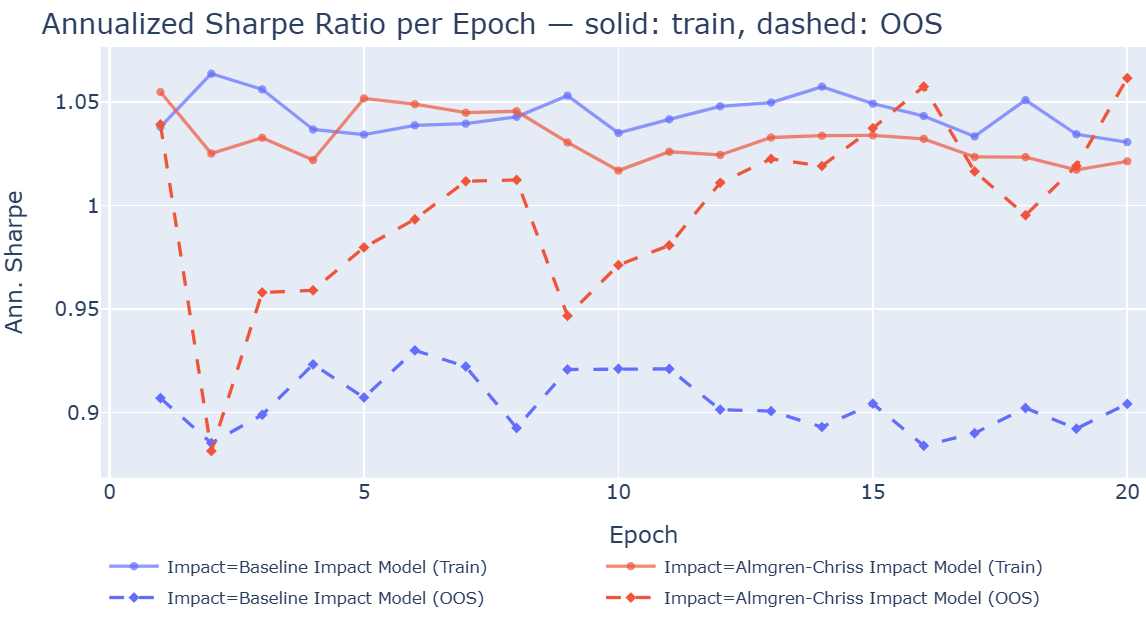}
\caption{TD3 (optimized) sharpe per epoch---POE, AC vs.\ 10\,bps baseline impact model.}
\label{fig:poe_td3_cost_model}
\end{figure}

Comparing TD3 (optimized) across cost models, the OOS convergence observed with the AC model does not occur with the 10\,bps baseline: baseline-impact OOS annualized return stays flat at $\sim$16\% while AC climbs to 19.35\%, and Sharpe remains at 0.9 for baseline vs.\ 1.06 for AC (Fig.~\ref{fig:poe_td3_cost_model}).
This improvement occurs despite higher turnover (1.77\% vs.\ 1.51\%) and POV (0.13 vs.\ 0.11) under AC, suggesting that the nonlinear cost signal teaches the agent a more generalizable policy.
A2C shows no meaningful difference between cost models.

\section{Limitations and Future Work}\label{sec:limitations}

Although our work improves many aspects of RL backtesting relative to existing environments, some were not addressed and can be improved in future work.

We use a fixed NASDAQ~100 composition; ideally the trading universe would reflect daily index membership, but this data is not freely available.
Expanding to the S\&P~500 or Russell~2000 would amplify cost differences, as participation rates in small-cap names are higher.

The AC model assumes linear permanent impact with constant coefficients; the Obizhaeva--Wang model~\cite{obizhaeva2013}, which captures transient impact decay, is implemented in our codebase but not yet systematically evaluated.
Carry costs (stock borrowing fees) are not calculated, which would be important for the margin trader.

The Optuna HPO pipeline currently optimizes for OOS Sharpe.
As shown in Section~\ref{sec:results}, optimized configurations can increase drawdowns while improving Sharpe, different optimization objectives might yield better overall results.

\section{Conclusion}\label{sec:conclusion}

We presented a suite of three Gymnasium-compatible trading environments with realistic, nonlinear market impact models and evaluated five DRL algorithms (A2C, PPO, DDPG, SAC, TD3) across stock trading, margin trading, and portfolio optimization formulations.
Our results demonstrate that:
(1)~the choice of cost model materially affects both absolute performance and the relative ranking of algorithms across all three environments---in stock trading, optimized PPO achieves 34\% OOS total return under the baseline but only 25\% under AC, while TD3 improves from 24\% to 30\%; in POE, TD3 under AC achieves 32\% total return (best overall) while the same algorithm under baseline yields only 26\%;
(2)~the AC model produces dramatically different trading behavior---non-optimized TD3 in stock trading sees daily costs drop from \$200k to \$8k (96\% reduction) when switching from baseline to AC, with turnover falling from 19\% to 1\%, while in POE the nonlinear cost model enables genuine training convergence that does not occur under flat fees;
(3)~hyperparameter optimization is essential not only for OOS performance but for constraining pathological execution---PPO and SAC exhibit rising POV and turnover across epochs without HPO (SAC stock trading costs drop 82\% with HPO; PPO POE IS POV reaches 1.39 without HPO vs.\ 0.36 with);
and (4)~algorithm choice interacts strongly with both the environment and cost model---DDPG's margin Sharpe jumps from $-$2.1 to 0.3 under AC while SAC's drops from $-$0.5 to $-$1.2, PPO is the best stock trading agent under baseline but is hurt by AC, and TD3 benefits most from AC across both stock and POE environments.

The full environment suite---with stock trading, margin trading, and portfolio optimization variants, plus Optuna HPO pipelines and comprehensive reporting---is released as an open-source extension to FinRL-Meta to support reproducible research in cost-aware reinforcement learning for finance.

\bibliographystyle{IEEEtran}
\bibliography{references}

\end{document}